\pdfoutput=1
\documentclass[11pt]{article}

\usepackage{acl}
\usepackage{float}
\usepackage{subfigure}
\usepackage{booktabs}
\usepackage{multirow}

\usepackage[normalem]{ulem}
\useunder{\uline}{\ul}{}
\usepackage{times}
\usepackage{diagbox}
\usepackage{amsmath}
\usepackage{amssymb}
\usepackage{latexsym}
\usepackage{graphicx}
\usepackage{algorithm}
\usepackage{algorithmic}

\usepackage{xcolor}
\usepackage{appendix}
\usepackage{paralist}

\usepackage[T1]{fontenc}
\usepackage[utf8]{inputenc}

\usepackage{microtype}

\title{MDERank: A Masked Document Embedding Rank Approach for Unsupervised Keyphrase Extraction}

\author{Linhan Zhang$^{12}$\thanks{~~Work is done during the internship at Speech Lab, Alibaba Group.} \quad Qian Chen$^{2}$ \quad Wen Wang$^{2}$ \quad Chong Deng$^{2}$ \quad Shiliang Zhang$^{2}$ \\ 
\textbf{\quad Bing Li$^{3}$ \quad Wei Wang$^{4}$ \quad Xin Cao$^{1}$}\\
$^1$School of Computer Science and Engineering, The University of New South Wales \\
$^2$Speech Lab, Alibaba Group, China\\
$^3$A*STAR Centre for Frontier AI Research (CFAR), Singapore \\
$^4$Hong Kong University of Science and Technology (Guangzhou), China \\
{\tt \{linahan.zhang, xin.cao\}@unsw.edu.au } \\
{\tt \{tanqing.cq,w.wang,dengchong.d,sly.zsl\}@alibaba-inc.com}\\
{\tt li\_bing@ihpc.a-star.edu.sg \quad  weiwcs@ust.hk}\\
}

\begin{document}
\maketitle
\begin{abstract}
Keyphrase extraction (KPE) automatically extracts phrases in a document that provide a concise summary of the core content, which benefits downstream information retrieval and NLP tasks. Previous state-of-the-art (SOTA) methods select candidate keyphrases based on the similarity between learned representations of the candidates and the document. They suffer performance degradation on long documents due to discrepancy between sequence lengths which causes mismatch between representations of keyphrase candidates and the document. In this work, we propose a novel unsupervised embedding-based KPE approach, Masked Document Embedding Rank (MDERank), to address this problem by leveraging a mask strategy and ranking candidates by the similarity between embeddings of the source document and the masked document. We further develop a KPE-oriented BERT (KPEBERT) model by proposing a novel self-supervised contrastive learning method, which is more compatible to MDERank than vanilla BERT. Comprehensive evaluations on six KPE benchmarks demonstrate that the proposed MDERank outperforms state-of-the-art unsupervised KPE approach by average 1.80 $F1@15$ improvement. MDERank further benefits from KPEBERT and overall achieves average 3.53 $F1@15$ improvement over the SOTA SIFRank. Our code is available at \url{https://github.com/LinhanZ/mderank}.
\end{abstract}

\section{Introduction}
\label{sec:intro}
Keyphrase extraction (KPE) automatically extracts a set of phrases in a document that provide a concise summary of the core content. KPE is highly beneficial for readers to quickly grasp the key information of a document and for numerous downstream tasks such as information retrieval and summarization. Previous KPE works include supervised and unsupervised approaches. Supervised approaches model KPE as sequence tagging \cite{sahrawat2019keyphrase, alzaidy2019bi,martinc2020tnt,santosh2020sasake,nikzad2021phraseformer} or sequence generation tasks \cite{liu2020keyphrase,kulkarni2021learning} and require large-scale annotated data to perform well. Since KPE annotations are expensive and large-scale KPE annotated data is scarce, unsupervised KPE approaches, such as TextRank~\cite{mihalcea2004textrank}, YAKE~\cite{campos2018yake}, EmbedRank~\cite{bennani2018simple}, are the mainstay in industry deployment.

Among unsupervised KPE approaches, embedding-based approaches including EmbedRank~\cite{bennani2018simple} and SIFRank~\cite{sun2020sifrank} yield the state-of-the-art (SOTA) performance. After selecting keyphrase (KP) candidates from a document using rule-based methods, embedding-based KPE approaches rank the candidates in a descending order based on a scoring function, which computes the similarity between embeddings of candidates and the source document. Then the top-$K$
candidates are chosen as the final KPs. We refer to these approaches as \emph{Phrase-Document-based} (PD) methods. 
PD methods have two major drawbacks: 

\begin{inparaenum}[(i)] 
\item As a document is typically significantly longer than candidate KPs and usually contains multiple KPs, it is challenging for PD methods to reliably measure their similarities in the latent semantic space. Hence, PD methods are naturally biased towards longer candidate KPs, as shown by the example in Table~\ref{table_case}.

\item The embedding of candidate KPs in the PD methods is computed without the contextual information, 
hence further limiting the effectiveness of the subsequent similarity match.
\end{inparaenum}

\begin{table*}[!htb]
\small
\centering
\begin{tabular}{l | p{11cm}}
 \toprule
Document &  The paper presents a method for pruning frequent itemsets based on background knowledge represented by a Bayesian network . The interestingness of an itemset is defined as the absolute difference between its support estimated from data and from the Bayesian network. Efficient algorithms are presented for finding interestingness of a collection of frequent itemsets , and  \ldots\\ \hline\hline
% for finding all attribute sets with a given minimum interestingness . Practical usefulness of the algorithms and their efficiency have been verified experimentally
% EmbedRank(BERT) 
SIFRank (Best PD method) & notation database attributes, research track paper dataset \#attrs max, bayesian network bn output, bayesian network computing, interactive network structure improvement process
% pruning frequent itemsets, secondary data mining problem, many interestingness measures, introduction finding frequent itemsets, interestingness measures %, interestingness measure, pruning methods, rule pruning, minimum interestingness, main drawback, active research area,  \textcolor{red}{ bayesian network}, practical usefulness, true probability distributions, hypothesis discovery process  
\\ \hline 
MDERank (Proposed method) & \textcolor{red}{interestingness}, pruning, \textcolor{red}{frequent itemsets}, pruning frequent itemsets, interestingness measures%, rules, \textcolor{red}{bayesian network} %, \textcolor{red}{association rules}, practical usefulness, methods, introduction finding frequent itemsets, pruning methods, secondary data mining problem, many interestingness measures,  rule pruning 
\\
\bottomrule
\end{tabular}
\caption{
An example shows the bias of Phrase-Document (PD) methods towards longer candidate keyphrases at $K=5$. 
%EmbedRank(BERT) is a modified version of EmbedRank, which replace Sent2Vec embeddings with BERT embeddings.  %
Keyphrase extracted are shown in a ranked order and those matching the gold labels are marked in red. }

\label{table_case}
\end{table*}

\begin{figure}[t]
\centering % 图片居中
\includegraphics[width=0.45\textwidth]{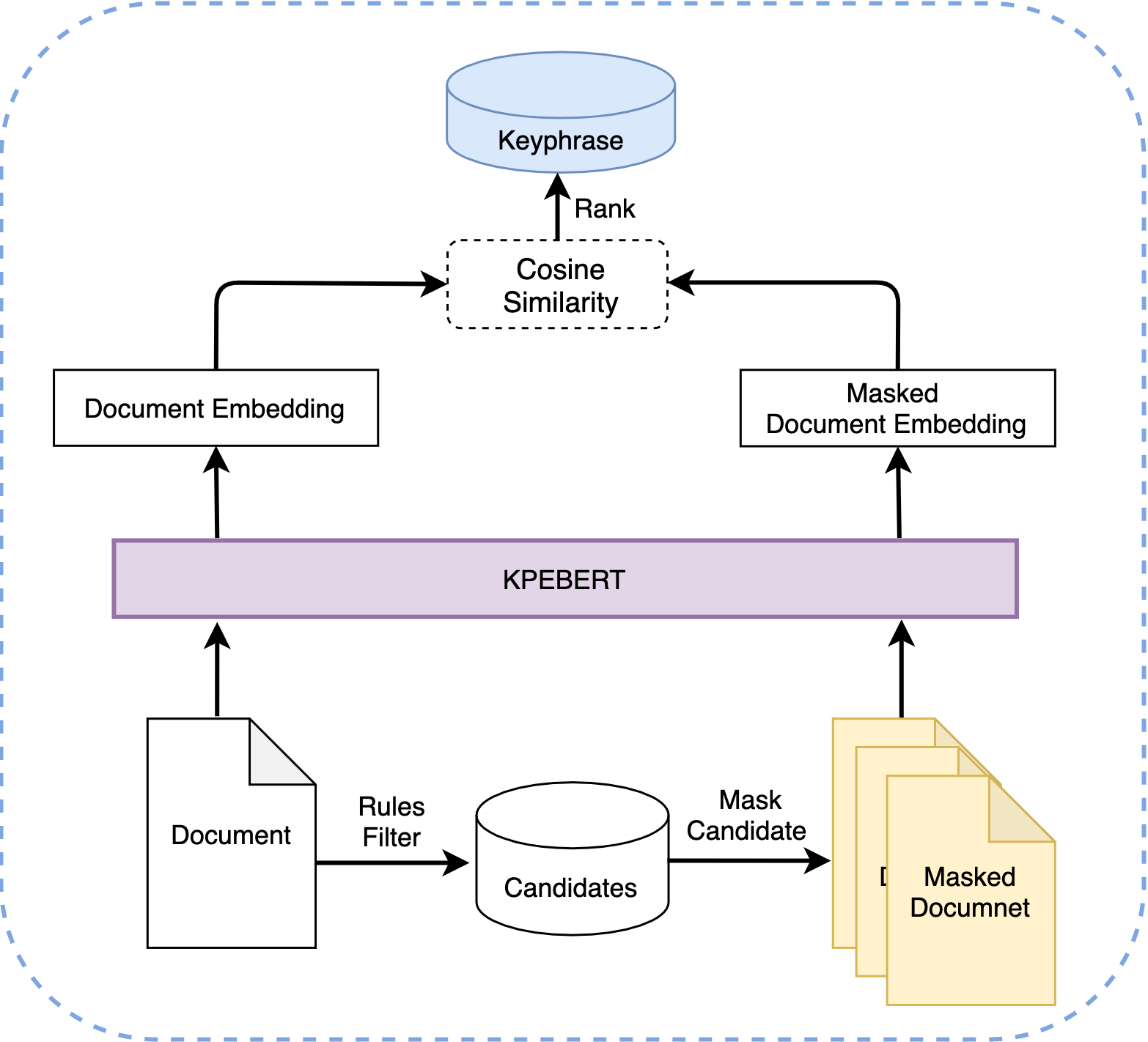}
\caption{
The architecture of the proposed MDERank approach. 
}
\label{figure_frame}
\end{figure}

In this paper, we propose a novel unsupervised embedding-based KPE method, denoted by Masked Document Embedding Rank (\textbf{MDERank}), to address above-mentioned drawbacks of PD methods. The architecture of MDERank is shown in Figure~\ref{figure_frame}.  The basic idea of MDERank is that a keyphrase plays an important role in the semantics of a document, and its absence from the document should cause a significant change in the semantics of the document.  Therefore, 
we propose to compare the embeddings of the original document and its variant where the occurrence(s) of some candidate KPs are masked.  
This leads to a new ranking principle based on the \emph{increasing} order of the resulting similarities, i.e., a \emph{lower} semantic similarity between the original document and its masked variant indicates a higher significance of the candidate. 

Our proposed method can be deemed as \emph{Document-Document} method and it addresses the two weaknesses of the \emph{Phrase-Document} methods:  
\begin{inparaenum}[(i)]
\item Since the sequence lengths of the original document and the masked document are the same, comparing their similarities in the semantic space is more meaningful and reliable. 
\item The embedding of the masked document is computed from a sufficient amount of context information and hence can capture the semantics reliably using the SOTA contextualized representation models such as BERT. 
\end{inparaenum} 
Inspired by~\citep{lewis2019bart,zhang2020pegasus,han2021ptr}, where pre-trained language models (PLMs) trained on objectives close to final downstream tasks achieve enhanced representations and improve fine-tune performance, 
we further propose a novel self-supervised contrastive learning method on top of BERT-based models (dubbed as \textbf{KPEBERT}). 

The main contributions of this work include:
\begin{itemize}
\item We propose a novel embedding-based unsupervised KPE approach (MDERank) that improves the reliability of computing KP candidate embeddings from contextualized representation models and improves robustness to different lengths of KPs and documents.
\item We propose a novel self-supervised contrastive learning method and develop a new pre-trained language model KPEBERT.
\item We conduct extensive evaluations of MDERank on six diverse KPE benchmarks and demonstrate the robustness of MDERank to different lengths of documents. MDERank with BERT achieves 17.00, 21.99, and 23.85 for average $F_1@5$, $F_1@10$, and $F_1@15$ respectively, as 1.69, 2.18 and 1.80 absolute gains over the SOTA results from SIFRank~\cite{sun2020sifrank}, and 4.44, 3.58, and 2.95 absolute gains over EmbedRank with BERT. MDERank with KPEBERT achieves further absolute gains by 1.70, 2.18 and 1.73.  Ablation analysis further provides insights into the effects of document lengths, encoder layers, and pooling methods.
\end{itemize}

\section{Related Work}
\label{sec:related}

\medskip
\noindent \textbf{Unsupervised KPE} 
Unsupervised KPE approaches do not require annotated data and there has been much effort in this line of research, as summarized in~\citep{papagiannopoulou2020review}. Unsupervised KPE approaches can be categorized into statistics-based, graph-based, and embedding-based methods. The statistics-based models such as YAKE~\cite{campos2018yake}, EQPM~\cite{li2017efficiently}, and CQMine~\cite{li2018efficient} explores both conventional position and frequency features and new statistical features capturing context information.
TextRank~\cite{mihalcea2004textrank} is a representative graph-based method, which converts a document into a graph based on lexical unit co-occurrences and applies PageRank iteratively. Many graph-based methods could be considered as modifications to TextRank by introducing extra features to compute weights for edges of the constructed graph, e.g., SingleRank~\cite{wan2008single}, PositionRank~\cite{florescu2017positionrank}, ExpandRank~\cite{wan2008single}. The graph-based TopicRank~\cite{bougouin2013topicrank} and MultipartiteRank~\cite{boudin2018unsupervised} methods enhance keyphrase diversity by constructing graphs based on clusters of candidate keyphrases. 
For embedding-based methods, \cite{wang2015using} first attempted on utilizing word embeddings as external knowledge base for keyphrases extraction and generation. Key2vec~\cite{mahata2018key2vec} used Fasttext to construct phrase/document embeddings and then apply PageRank to select keyphrases from candidates. EmbedRank~\cite{bennani2018simple} measures the similarity between phrase and document embeddings for ranking. SIFRank~\cite{sun2020sifrank} improves the static embeddings in EmbedRank by a pre-trained language model ELMo and a sentence embedding model SIF~\cite{arora2016simple}. KeyBERT\footnote{\url{https://maartengr.github.io/KeyBERT/}} is a tooltik for keyphrase extraction with BERT, following the PD based methods paradigm. AttentionRank~\cite{ding2021attentionrank} used a pretrained language model to calculate self-attention of a candidate within the context of a sentence and cross-attention between a candidate and sentences within a document, in order to evaluate the local and global importance of candidates. As analyzed in Section~\ref{sec:intro}, for embedding-based methods, using contextualized embedding models to compute candidate embeddings could be unreliable due to lack of context, and these methods lack robustness to different lengths of keyphrases and documents. Our proposed MDERank approach could effectively address these drawbacks.

\noindent \textbf{Contextual Embedding Models} 
Early emebdding models include static word embeddings such as Word2Vec~\cite{mikolov2013efficient}, GloVe~\cite{pennington2014glove}, and FastText~\cite{bojanowski2017enriching}, phrase embedding model HCPE~\cite{li2018adaptive}, and sentence embeddings such as Sent2Vec~\cite{pagliardini2017unsupervised} and Doc2Vec~\cite{lau2016empirical}, which render word or sentence representations that do not depend on their context. In contrast, pre-trained contextual embedding models, such as ELMo~\cite{peters2018deep}, incorporate rich syntactic and semantic information from context for representation learning and yield more expressive representations. BERT~\cite{devlin2018bert} captures better context information through a bidirectional transformer encoder than the Bi-LSTM based ELMo, and has established SOTA in a wide variety of NLP tasks. In one line of research, RoBERTa~\cite{liu2019roberta}, XLNET~\cite{yang2019xlnet} and many other BERT variant PLMs have been proposed to further improve the language representation capability. In another line of research, Longformer~\cite{beltagy2020longformer}, BigBird~\cite{zaheer2020big} and other efficient transformers are proposed to reduce the quadratic complexity of transformer on sequence length in order to model long-range dependencies. In this paper, we mainly use BERT as the default contextual embedding model. We also evaluate the performance of MDERank with these efficient transformers on long documents.

\begin{figure*}[!htb]
\centering
\includegraphics[width=0.7\textwidth]{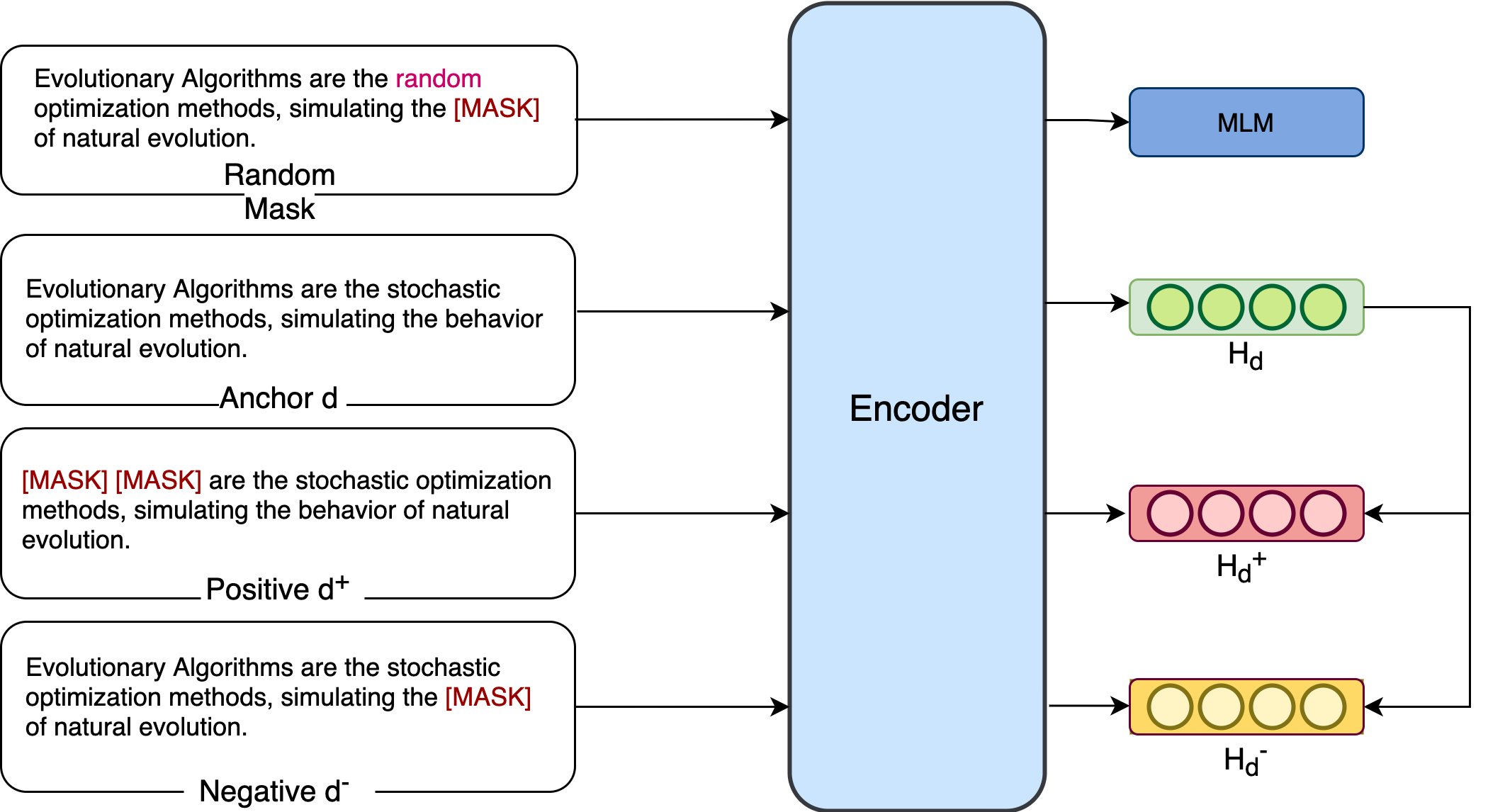}
\caption{The multi-task pre-training for KPEBERT includes two tasks. One is teaching the encoder to distinguish documents masked with keyphrases and non-keyphrases. The other is further pre-training the encoder with a MLM task. The word in pink is an example to illustrate the random masking in MLM.}
\label{figure_mlm}
\end{figure*}

\section{MDERank}
\label{sec:mderank}
In this section, we describe the proposed Masked Document Embedding Rank (MDERank) approach.
%\section{Problem Definition}
Given a document $d = \{w_1, w_2, \ldots, w_n\}, d \in D$ where $D$ denotes a dataset,  and a set of selected candidate KPs  $C=\{c_1, \ldots, c_i, \ldots, c_m\}$, where a candidate $c_i$ consists of one or multiple tokens, as $c_i = \{c_i^1, \ldots,c_i^l\}$, and  $m\leq n$,  KPE aims to select $K$ candidates from $C$, where $K \leq m$. Following the common practice \cite{bennani2018simple,sun2020sifrank}, after tokenization and POS tagging, candidates are selected with continuous regular expression \verb!<NN. *|JJ> * <NN.*>!, which are mostly noun phrases.

To address the mismatch between sequence lengths of a document and a candidate phrase as well as lack of contextual information in PD methods as mentioned in Section \ref{sec:intro}, we hypothesize that it is more reasonable to conduct the similarity comparison at the \emph{document-document} level rather than at the \emph{phrase-document} level. 

Based on this hypothesis, for each candidate KP $c_i$ for a document $d$, given its occurrence positions in $d$ as $[p_1, p_2, \ldots, p_t]$, MDERank replaces all occurrences of $p_{i=1}^t$ by a special placeholder token \verb|MASK|. It is noted the number of \verb|MASK| used for masking $p_t$ is as same as the number of tokens in $c_i$. And then we construct a masked variant of the original document as $d_{M}^{c_i}$.  We define the similarity score $f(c_i)$ for ranking the significance of candidates as the cosine similarity between $E(d)$ and $E(d_M^{c_i})$, where $E(\cdot)$ represents the embedding of a document.  Note that a higher $f(c_i)$ value indicates a lower ranking for $c_i$, which is opposite to the PD methods. This is because the higher similarity the less important the candidate $c_i$ is. The semantic of masked document is not changed much compared with original one as only a trivial phrase is masked. We use BERT~\cite{devlin2018bert} as the default embedding model and investigate other contextual embedding models in Section~\ref{sec:analyses}.
BERT is pre-trained through self-supervised tasks of masked language modeling (MLM) and next sentence prediction (NSP), on large-scale unlabeled text of English Wikipedia (2500M words) and Bookscorpus (800 words). A document $d = \{w_1, w_2, \ldots, w_n\}$ is prepended with a special token \verb![CLS]! and then encoded by BERT to obtain the hidden representations of tokens as $\{H_1, H_2, \ldots, H_n\}$.  The document embedding $E(d)$ is computed as follows:
\begin{equation}
\label{eq:doc-embedding}
    E(d) = \textrm{MaxPool}(H_1,\ldots,H_n)
\end{equation}
We also investigate average pooling in Section~\ref{sec:analyses} and other masking methods in Appendix~\ref{appendix:masking}.

\begin{table}[h]
\centering
\scalebox{0.9}{%
\begin{tabular}{@{}c|ccc@{}}
\toprule
\textbf{Datasets} & \textbf{$N_{KP}$} & \textbf{$L_{KP}$} & \textbf{$L_{Doc}$} \\ \midrule
Inspec      & 9.82  & 2.31 & 121.84  \\
SemEval2010 & 15.07 & 2.11 & 189.90  \\
SemEval2017 & 17.30 & 3.00 & 170.38  \\
DUC2001     & 8.08  & 2.07 & 724.63  \\
NUS         & 11.66 & 2.07 & 7702.00    \\
Krapivin    & 5.74  & 2.03 & 8544.57 \\ \bottomrule
\end{tabular}%
}
\caption{Statistics of the datasets. $N_{KP}$ is the average number of gold keyphrases. $L_{KP}$ is the average length of gold keyphrases. $L_{Doc}$ is the average number of words per document.}
\label{table_dataset}
\end{table}

\section{KPEBERT: KPE-oriented Self-supervised Learning}%
\label{sec:kpebert}
BERT and many other BERT variant PLMs can effectively capture syntactic and semantic information in language representations for downstream NLP tasks, through self-supervised learning objectives such as MLM. However, these self-supervised learning objectives neither explicitly model the significance of KPs nor model ranking between KPs. In this paper, we propose a novel PLM KPEBERT trained with a novel self-supervised learning objective to improve the capabilities of PLMs for ranking KPs. %
This new task is defined as minimizing the triplet loss between positive and negative examples (See Figure~\ref{figure_mlm}). After obtaining a set of pseudo-KPs for documents using another unsupervised KPE method $\theta$, we define documents masking out pseudo-KPs as positive examples while those masking out ``non-pseudo-KPs'' as negative examples. Following SimCSE~\cite{gao2021simcse}, we encode the original document $d$ (anchor), the positive example $d^+$, and negative example $d^-$ through a PLM encoder, respectively, and obtain their hidden representations as $H_d$, $H_{d^+}$, and $H_{d^-}$. 

Finally, we define the triplet loss as:
\begin{equation}
\begin{split}
\ell_{CL} =& \max(sim(H_d,H_{d^+}) \\
& - sim(H_d,H_{d^-}) + m, 0)
\end{split}
\end{equation}

\noindent where $sim(H_x,H_y)$ denotes the similarity between embeddings of the document $x$ and $y$. We use cosine similarity (same as used for MDERank). $m$ is a margin parameter. %

We initialize KPEBERT from BERT-base-uncased\footnote{https://huggingface.co/bert-base-uncased} and then incorporate the standard MLM pre-training task as in BERT into the overall learning objective to avoid forgetting the previously learned general linguistic knowledge, as follows:
\begin{equation}
\label{eq:all_loss}
    \ell = \ell_{CL} + \lambda \cdot \ell_{MLM}
\end{equation}

where $\lambda$ is a hyper-parameter balancing the two losses in the multi-task learning setting. KPEBERT differs from SimSCE in two major aspects: \begin{inparaenum}[(i)] 
\item  KPEBERT uses pseudo labeling and positive/negative example sampling strategies (below), different from standard dropout used by SimCSE to construct pair examples;
\item KPEBERT uses triplet loss whereas SimCSE uses contrastive loss. 
% \item KPEBERT adds MLM loss to CL loss in multi-task learning whereas SimCSE only uses CL loss.
\end{inparaenum}

\begin{table*}[!htb]
\centering
\resizebox{\textwidth}{!}{%
\begin{tabular}{@{}c|c|cccccc|c|c@{}}
\toprule
\multirow{2}{*}{$F_1@K$} &
  \multirow{2}{*}{\textbf{Method}} &
  \multicolumn{6}{c|}{\textbf{Dataset}} &
  \multirow{2}{*}{\textbf{AVG}} &
  \multirow{2}{*}{\textbf{AvgRank (STD)}} \\ \cmidrule(lr){3-8}
 &
   &
  \textbf{Inspec} &
  \textbf{SemEval2017} &
  \textbf{SemEval2010} &
  \textbf{DUC2001} &
  \textbf{Krapivin} &
  \textbf{NUS} &
   &
   \\ \midrule
\multirow{13}{*}{5} &
  TextRank &
  21.58 &
  16.43 &
  7.42 &
  11.02 &
  6.04 &
  1.80 &
  10.72 &
  9.33 $(\pm{1.60})$ \\
 &
  SingleRank &
  14.88 &
  18.23 &
  8.69 &
  19.14 &
  8.12 &
  2.98 &
  12.01 &
  7.67 $(\pm{0.94})$ \\
 &
  TopicRank &
  12.20 &
  17.10 &
  9.93 &
  19.97 &
  8.94 &
  4.54 &
  12.11 &
  7.17 $(\pm{1.77})$ \\
 &
  MultipartiteRank &
  13.41 &
  17.39 &
  10.13 &
  21.70 &
  9.29 &
  6.17 &
  13.02 &
  6.17 $({\pm{1.77}})$ \\
 &
  YAKE &
  8.02 &
  11.84 &
  6.82 &
  11.99 &
  8.09 &
  7.85 &
  9.10 &
  9.00 $(\pm{2.52})$ \\
 &
  EmbedRank(Sent2Vec)+MMR &
  14.51 &
  20.21 &
  9.63 &
  21.75 &
  8.44 &
  2.13 &
  12.78 &
  6.83 $(\pm{2.03})$ \\
 &
  SIFRank(ELMo) &
  {\ul \textbf{29.38}} &
  \textbf{22.38} &
  11.16 &
  {\ul \textbf{24.30}} &
  1.62 &
  3.01 &
  15.31 &
  4.50 $(\pm{3.77})$ \\
 &
  EmbedRank(BERT) &
  28.92 &
  20.03 &
  10.46 &
  8.12 &
  4.05 &
  3.75 &
  12.56 &
  6.83 $(\pm{3.02})$ \\ \cmidrule(lr){2-10}
 &
  MDERank(BERT) &
  26.17 &
  {\ul \textbf{22.81}} &
  \textbf{12.95} &
  13.05 &
  11.78 &
  {\ul \textbf{15.24}} &
  17.00 &
  3.33 $(\pm{2.49})$ \\
 &
  MDERank(KPEBERT$_{ab}$) &
  28.06 &
  21.63 &
  \textbf{12.95} &
  22.51 &
  {\ul \textbf{12.91}} &
  14.11 &
  {\ul \textbf{18.70}} &
  {\textbf{2.67}} $(\pm{0.75})$ \\
 &
  MDERank(KPEBERT$_{re}$) &
  27.85 &
  20.37 &
  {\ul \textbf{13.05}} &
  \textbf{23.31} &
  \textbf{12.35} &
  \textbf{14.39} &
  \textbf{18.55} &
  {\ul \textbf{2.50}} $(\pm{1.12})$ \\ \midrule \midrule
\multirow{13}{*}{10} &
  TextRank &
  27.53 &
  25.83 &
  11.27 &
  17.45 &
  9.43 &
  3.02 &
  15.76 &
  8.00 $(\pm{1.63})$\\
 &
  SingleRank &
  21.50 &
  27.73 &
  12.94 &
  23.86 &
  10.53 &
  4.51 &
  16.85 &
  6.67 $(\pm{1.49})$\\
 &
  TopicRank &
  17.24 &
  22.62 &
  12.52 &
  21.73 &
  9.01 &
  7.93 &
  15.18 &
  8.50 $(\pm{1.50})$\\
 &
  MultipartiteRank &
  18.18 &
  23.73 &
  12.91 &
  24.10 &
  9.35 &
  8.57 &
  16.14 &
  7.17 $(\pm{1.67})$ \\
 &
  YAKE &
  11.47 &
  18.14 &
  11.01 &
  14.18 &
  9.35 &
  11.05 &
  12.53 &
  9.17 $(\pm{2.54})$ \\
 &
  EmbedRank(Sent2Vec)+MMR &
  21.02 &
  29.59 &
  13.9 &
  25.09 &
  10.47 &
  2.94 &
  17.17 &
  6.67 $(\pm{2.29})$ \\
 &
  SIFRank(ELMo) &
  {\ul \textbf{39.12}} &
  {\ul \textbf{32.60}} &
  16.03 &
  {\ul \textbf{27.60}} &
  2.52 &
  5.34 &
  20.54 &
  4.50 $(\pm{3.91})$\\
 &
  EmbedRank(BERT) &
  \textbf{38.55} &
  31.01 &
  16.35 &
  11.62 &
  6.60 &
  6.34 &
  18.41 &
  6.50 $(\pm{3.20})$ \\ \cmidrule(lr){2-10}
 &
  MDERank(BERT) &
  33.81 &
  \textbf{32.51} &
  17.07 &
  17.31 &
  12.93 &
  \textbf{18.33} &
  21.99 &
  4.00 $(\pm{2.45})$ \\
 &
  MDERank(KPEBERT$_{ab}$) &
  35.80 &
  32.23 &
  \textbf{17.95} &
  \textbf{26.97} &
  {\ul \textbf{14.36}} &
  17.72 &
  {\ul \textbf{24.17}} &
  {\ul \textbf{2.33}} $(\pm{0.75})$ \\
 &
  MDERank(KPEBERT$_{re}$) &
  34.36 &
  31.21 &
  {\ul \textbf{18.27}} &
  26.65 &
  \textbf{14.31} &
  {\ul \textbf{18.46}} &
  \textbf{23.88} &
  \textbf{2.50} $(\pm{1.26})$ \\ \midrule \midrule
\multirow{13}{*}{15} &
  TextRank &
  27.62 &
  30.50 &
  13.47 &
  18.84 &
  9.95 &
  3.53 &
  17.32 &
  8.00 $(\pm{1.73})$\\
 &
  SingleRank &
  24.13 &
  31.73 &
  14.4 &
  23.43 &
  10.42 &
  4.92 &
  18.17 &
  6.67 $(\pm{1.49})$ \\
 &
  TopicRank &
  19.33 &
  24.87 &
  12.26 &
  20.97 &
  8.30 &
  9.37 &
  15.85 &
  8.83 $(\pm{1.77})$ \\
 &
  MultipartiteRank &
  20.52 &
  26.87 &
  13.24 &
  23.62 &
  9.16 &
  10.82 &
  17.37 &
  7.33 $(\pm{1.80})$ \\
 &
  YAKE &
  13.65 &
  20.55 &
  12.55 &
  14.28 &
  9.12 &
  13.09 &
  13.87 &
  9.00 $(\pm{2.45})$ \\
 &
  EmbedRank(Sent2Vec)+MMR &
  23.79 &
  33.94 &
  14.79 &
  24.68 &
  10.17 &
  3.56 &
  18.49 &
  6.50 $(\pm{1.98})$ \\
 &
  SIFRank(ELMo) &
  {\ul \textbf{39.82}} &
  \textbf{37.25} &
  18.42 &
  {\ul \textbf{27.96}} &
  3.00 &
  5.86 &
  22.05 &
  4.67 $(\pm{3.77})$\\
 &
  EmbedRank(BERT) &
  \textbf{39.77} &
  36.72 &
  19.35 &
  13.58 &
  7.84 &
  8.11 &
  20.90 &
  6.33 $(\pm{3.30})$ \\ \cmidrule(lr){2-10}
 &
  MDERank(BERT) &
  36.17 &
  37.18 &
  20.09 &
  19.13 &
  12.58 &
  \textbf{17.95} &
  23.85 &
  4.00 $(\pm{2.00})$ \\
 &
  MDERank(KPEBERT$_{ab}$) &
  37.43 &
  {\ul \textbf{37.52}} &
  {\ul \textbf{20.69}} &
  26.28 &
  {\ul \textbf{13.58}} &
  \textbf{17.95} &
  {\ul \textbf{25.58}} &
  {\ul \textbf{2.00}} $(\pm{1.00})$ \\
 &
  MDERank(KPEBERT$_{re}$) &
  36.40 &
  36.63 &
  {\textbf{20.35}} &
  {\textbf{26.42}} &
  {\textbf{13.31}} &
  {\ul \textbf{19.41}} &
  \textbf{25.42} &
  \textbf{2.67} $(\pm{1.37})$ \\ \bottomrule
\end{tabular}%
}
\caption{
\small{
 KPE performance as $F_1@K, K \in \{5,10,15\}$ on the six benchmarks. For each $K$, the first group shows performance of the baselines and the second group shows performance of our proposed MDERank using BERT for embedding and MDERank using KPEBERT for embedding. EmbedRank(BERT) denotes the Phrase-Document based methods for a fair comparison. The best results are both underlined and in bold. The second-best results are in bold. \textbf{Ab} and \textbf{Re} denote absolute and relative sampling, respectively.  \textbf{AVG} is the average $F_1@K$ on all six benchmarks. \textbf{AvgRank(STD)} is the mean and std of the rank of a method among all methods on all six benchmarks (hence the lower the better).}
}
\label{table_main}
\end{table*}
\paragraph{Absolute Sampling} %
For a document $d$, we first select candidate keyphrases $C$ using POS tags with regular expressions as described in Section~\ref{sec:mderank}. Then we obtain a set of keyphrases $C'$ extracted by another unsupervised KPE approach $\theta$ on $d$, as pseudo labels. We define ``keyphrases'' as $C'$ and ``non-keyphrases'' as $C \setminus C'$. We mask a ``keyphrase'' from a document with a \verb|MASK| to construct a positive example $d^+$ for $d$. We select a ``non-keyphrase'' and perform the same mask operation to construct a negative example $d^-$.

\paragraph{Relative Sampling} %
In this approach, after obtaining a set of KP $C'$ extracted by $\theta$, 
we randomly select a pair of KPs from $C'$ and choose the one ranked higher to construct a positive example and the other one to construct a negative example through the mask operation. On one hand, the decisions of ``keyphrases'' and ``non-keyphrases'' are fully based on the ranking predicted by $\theta$, hence relative sampling may increase the impact from $\theta$ on the inductive bias of KPEBERT. On the other hand, relative sampling mines more hard negative samples which may improve performance of triplet loss based learning. We study the efficacy of these two sampling approaches on KPEBERT in Section~\ref{sec:exp-f1-cmp}.

\section{Experiments}
\label{sec:expt}
\subsection{Datasets and Metrics}
The pre-training data for KPEBERT are the WikiText language modeling dataset with 100+ million tokens extracted from a set of verified Good and Featured articles on Wikipedia\footnote{\url{https://huggingface.co/datasets/wikitext}}. We use six KPE benchmarks for evaluations. Four of them are scientific publications\footnote{\url{https://github.com/memray/OpenNMT-kpg-release}}, including Inspec~\cite{hulth2003improved}, Krapivin~\cite{krapivin2009large}, NUS~\cite{nguyen2007keyphrase}, and SemEval-2010~\cite{kim2010semeval}, all widely used for evaluations in previous works~\cite{meng2017deep,chen2019integrated,sahrawat2019keyphrase,bennani2018simple,meng2020empirical}. We also evaluate on the DUC2001 dataset (news articles)~\cite{wan2008single} and SemEval2017 dataset (science journals)~\cite{augenstein2017semeval}\footnote{\url{https://github.com/sunyilgdx/SIFRank/tree/master/data}}. Table~\ref{table_dataset} shows data statistics. For a fair comparison with SIFRank, we use the entire documents, including abstract and main body.
%These datasets can be categorized into short-document, medium-length-document, and long-document datasets according to the average number of words per document.
Following previous works, predicted KPs are deduplicated and the KPE performance is evaluated as F$_1$ at the top K KPs ($K \in \{5, 10, 15\}$). Stemming is applied for computing F$_1$.

\subsection{Baselines and Training Details} %
\label{sec:baseline}
The first group for each $K$ in Table~\ref{table_main} shows performance of the eight baseline unsupervised KPE approaches. We evaluate TextRank, SingleRank, TopicRank, MultipartiteRank, YAKE, EmbedRank using their implementations in the widely used toolkit PKE\footnote{\url{https://github.com/boudinfl/pke}} with the default parameter settings. We evaluate SIFRank using the codebase \footnote{\url{https://github.com/sunyilgdx/SIFRank/tree/master}} and the same parameters suggested by the authors~\cite{sun2020sifrank}. The original EmbedRank~\cite{bennani2018simple} uses Sent2Vec for embedding and introduces embedding-based maximal marginal relevance (MMR) for improving diversity among selected KPs. For a fair comparison between the Phrase-Document method and our Document-Document MDERank, we design a new baseline EmbedRank(BERT) by replacing Sent2Vec with BERT and removing MMR. Some previous works have inflated results caused by ignoring deduplication and stemming, which are not fair in practice. Therefore, SIFRank and EmbedRank, which exclude such biases, are strong baselines for unsupervised keyphrase extraction with SIFRank considered to be the previous SOTA.

%The baselines include six unsupervised KPE approaches implemented in the widely used toolkit PKE\footnote{\url{https://github.com/boudinfl/pke}}, the SIFRank\footnote{\url{https://github.com/sunyilgdx/SIFRank/tree/master}}, and EmbedRank(BERT). Either the default parameter settings or those suggested by authors were used. %

%The EmbedRank(BERT) is specially designed for a fair comparison of the Phrase-Document approach and our Document-Document approach, using the same BERT model. We modify the EmbedRank implementation in PKE to implement EmbedRank(BERT), by replacing Sent2Vec with BERT embeddings and removing the original mechanisms for improving diversity among selected KPs.

%the diversity handling part. 
%(MMR) in EmbedRank. 
%For all baselines, we use Stanford CoreNLP\footnote{\url{https://stanfordnlp.github.io/CoreNLP/}} to tokenize and generate POS tags. 
% Regular expression \verb!<NN. *|JJ> * <NN.*>! is used to extract noun phrases as candidates.

%% Effects of the choice of $\theta$ on KPEBERT are analyzed in Appendix~\ref{sec:analyses} where we compare YAKE and TextRank as $\theta$.

We use YAKE~\cite{campos2018yake} as $\theta$ to extract ``keyphrases'' for a document for KPEBERT pre-training, due to its high efficiency and consistent performance. Effects of the choice of $\theta$ on KPEBERT are analyzed in Section~\ref{sec:analyses} where we compare YAKE and TextRank as $\theta$. The number of pseudo labels for absolute and relative sampling for KPEBERT pre-training are 10 and 20, respectively. The $\lambda$ is set to 0.1. The default parameter setting is the same as~\cite{gao2021simcse} except that we set the margin $m$ for triplet loss to 0.2 and the learning rate to 3e-5. We use 4 NVIDIA V100 GPUs for training, the batch size is 2 per device and the gradient accumulation is 4. We train 10 epochs.

\subsection{Performance Comparison}
\label{sec:exp-f1-cmp}
Table~\ref{table_main} shows F$_1$ at the top $K \in \{5, 10, 15\}$ predictions. For each $K$,  the first group shows the baseline results, and the second group shows results from our MDERank(BERT) (default using BERT for embedding) and MDERank using KPEBERT for embedding, MDERank(KPEBERT). MDERank(BERT) and MDERank(KPEBERT) perform consistently well on all benchmarks. MDERank(BERT) outperforms EmbedRank(BERT) by 2.95 average $F_1@15$ and outperforms the previous SOTA SIFRank by 1.80 average $F_1@15$. MDERank further benefits from KPEBERT and MDERank(KPEBERT) achieves 3.53 average $F_1@15$ gain over SIFRank,
%and outperform SIFRank and EmbedRank, 
especially on long-document datasets NUS and Krapivin. We also compute the average recalls of KPs with different phrase lengths (PL) in top-15 extracted KPs on all 6 benchmarks, for both EmbedRank(BERT) and MDERank(BERT), as shown in Table~\ref{table_wn}. We observe that EmbedRank has a strong bias for longer phrases, with PLs of its extracted KPs concentrated in [2,3]; whereas, PLs of KPs extracted by MDERank are more evenly distributed on diverse datasets. This analysis confirms that MDERank indeed alleviates the bias towards longer phrases from EmbedRank.

However, we observe that MDERank(BERT) has a large gap to SIFRank on DUC2001 and performs worse than EmbedRank(BERT) on Inspec. We investigate the reasons for these poorer performance. Different from other datasets collected from scientific publications, DUC2001 consists of open-domain news articles. The previous SOTA SIFRank introduces domain adaptation by combining weights from common corpus and domain corpus in the weight function of words for computing sentence embeddings, which may contribute significantly to its superior performance on DUC2001. As the default embedding model for MDERank, BERT is pre-trained on open-domain Wikipedia and BooksCorpus. However, as explained in Section~\ref{sec:kpebert}, BERT does not emphasize learning significance of KPs or ranking between KPs. KPEBERT is designed to tackle this problem. Although the training data for KPEBERT, the open-domain WikiText language modeling dataset, is much smaller than English Wikipedia, with KPE-oriented representation learning in KPEBERT, the performance of MDERank(KPEBERT) improves remarkably and is comparable to SIFRank. For Inspec, the average PLs of gold labels of this dataset is relatively high (see Table~\ref{table_dataset}). Also, on this dataset, when we move candidates with only 1 token to the end of ranking, MDERank(BERT) improves $F_1@5$, $F_1@10$, $F_1@15$ to 29.71 ,38.15, 39.46, an improvement of 3.54, 4.34 and 3.29, respectively. These analyses show that gold labels for Inspec are biased towards long PL. Therefore, EmbedRank with inductive bias for long PL may benefit from this annotation bias and perform well. However, MDERank still outperforms baselines based on its best average F$_1$ and top average rank among all methods on all datasets, proving its robustness across domains without any domain adaptation.

It is notable that MDERank particularly outperforms baselines on long-document datasets, verifying that MDERank could mitigate the weakness of performance degradation on long documents from PD methods. We further investigate effects of document length in Section~\ref{sec:analyses}. %
Absolute and relative sampling for KPEBERT achieve comparable performance on the 6 benchmarks with absolute sampling gaining a very small margin. Relative sampling performs better on NUS but is worse on Inspec and SemEval2017. %
We plan to continue exploring  sampling approaches in future work, to reduce dependency on $\theta$ and improve KPEBERT.

\begin{table}[t]
\centering
\scalebox{0.6}{
\begin{tabular}{@{}c|cccc|cccc@{}}
\toprule
\textbf{Method}  & \multicolumn{4}{c|}{\textbf{EmbedRank(BERT)}} & \multicolumn{4}{c}{\textbf{MDERank(BERT)}}    \\ \midrule
\diagbox{Data}{PL} & 1       & 2       & 3      & \textgreater{}3  & 1     & 2     & 3     & \textgreater{}3
\\ \midrule

Inspec            & 24.80   & 54.53   & 46.11  & 21.57            & 27.90 & 48.71 & 43.20 & 21.17           \\
SemEval2017       & 24.91   & 53.68   & 48.05  & 9.84             & 37.28 & 47.07 & 43.99 & 9.76            \\
SemEval2010       & 9.35    & 22.79   & 18.07  & 4.17             & 21.55 & 19.99 & 15.95 & 4.17            \\
DUC2001           & 3.46    & 19.39   & 37.39  & 15.58            & 24.81 & 23.70 & 23.66 & 13.46           \\
Krapivin          & 4.31    & 13.59   & 11.80  & 2.50             & 15.88 & 22.43 & 10.62 & 2.14            \\
NUS               & 5.12    & 9.53    & 16.17  & 2.84             & 26.77 & 24.70 & 17.12 & 1.90            \\ \bottomrule
\end{tabular}%
}
\caption{The average recall of predicted KPs with different phrase lengths (PL) on all six benchmarks, from EmbedRank(BERT) and MDERank(BERT).
}
\label{table_wn}
\end{table}

\subsection{Analyses}
\label{sec:analyses}

%\subsection{Effects of Document Length}\label{sec:exp-doc-len} 
\textbf{Effects of Document Length}
Section~\ref{sec:exp-f1-cmp} demonstrates the superior performance of MDERank especially on long documents. We conduct two experiments to further analyze effects of document length on the performance of PD methods and MDERank. We choose EmbedRank(BERT) to represent PD methods. In the first experiment, both approaches use BERT for embedding and we truncate a document into the first 128, 256, 512 words. As shown in Table~\ref{table_length}, F$_1$ for EmbedRank(BERT) drops drastically as the document length increases from 128 to 512, reflecting the weakness of EmbedRank(BERT) that the increased document length exacerbates discrepancy between sequence lengths of the document and KP candidates and mismatches between their embeddings, which degrades the KPE performance. In contrast, the performance of MDERank(BERT) improves steadily with increased document lengths, demonstrating the robustness of MDERank to document lengths and its capability to improve KPE from more context in longer documents.

\begin{table}[]
\centering
\scalebox{0.75}{%
\begin{tabular}{@{}c|c|ccc@{}}
\toprule
\textbf{Method}                  & \textbf{Doc Len} & \textbf{F$_1$@5} & \textbf{F$_1$@10} & \textbf{F$_1$@15} \\ \midrule
\multirow{3}{*}{EmbedRank(BERT)} & 128      & 8.76          & 14.75          & 16.28          \\
                                 & 256      & 5.86          & 10.19          & 12.90          \\
                                 & 512      & 3.75          & 6.34           & 8.11           \\ \midrule
\multirow{3}{*}{MDERank(BERT)}         & 128      & 12.86         & 16.06          & 16.67          \\
                                 & 256      & 14.45         & 16.01          & 16.64          \\
                                 & 512      & 15.24         & 18.33          & 17.95          \\ \bottomrule
\end{tabular}%
}
\caption{Effects of document lengths (the first 128, 256, 512 words of a document) on the KPE performance on the NUS dataset from EmbedRank(BERT) and MDERank(BERT).}
\label{table_length}
\end{table}

\begin{table}[]
\centering
\scalebox{0.65}{%
\begin{tabular}{@{}c|l|c|ccc@{}}
\toprule
\multirow{2}{*}{\textbf{Method}} &
  \multicolumn{1}{c|}{\multirow{2}{*}{\textbf{Pooling}}} &
  \multirow{2}{*}{\textbf{Layer}} &
  \multicolumn{3}{c}{\textbf{DUC2001}} \\ \cmidrule(l){4-6} 
 &
  \multicolumn{1}{c|}{} &
   &
  \textbf{F1@5} &
  \textbf{F1@10} &
  \textbf{F1@15} \\ \midrule
\multirow{6}{*}{EmbedRank(BERT)} & \multirow{3}{*}{AvgPooling}                      & 3  & 16.19 & 21.21 & 22.12 \\
                                 &                                                  & 6  & 10.76 & 15.33 & 17.63 \\
                                 &                                                  & 12 & 10.41 & 15.15 & 17.69 \\ \cmidrule(l){2-6} 
                                 & \multirow{3}{*}{MaxPooling}                      & 3  & 6.97  & 11.04 & 12.27 \\
                                 &                                                  & 6  & 7.12  & 10.93 & 13.13 \\
                                 &                                                  & 12 & 8.12  & 11.62 & 13.58 \\ \midrule
\multirow{6}{*}{MDERank(BERT)}   & \multirow{3}{*}{AvgPooling}                      & 3  & 12.00 & 16.45 & 19.08 \\
                                 &                                                  & 6  & 12.40 & 17.07 & 19.02 \\
                                 &                                                  & 12 & 13.00 & 17.93 & 19.45 \\ \cmidrule(l){2-6} 
                                 & \multicolumn{1}{c|}{\multirow{3}{*}{MaxPooling}} & 3  & 11.06 & 16.16 & 18.01 \\
                                 & \multicolumn{1}{c|}{}                            & 6  & 11.06 & 15.91 & 17.98 \\
                                 & \multicolumn{1}{c|}{}                            & 12 & 13.05 & 17.31 & 19.13 \\ \bottomrule
\end{tabular}%
}
\caption{KPE performance  on DUC2001 from EmbedRank(BERT) and MDERank(BERT) using different BERT layers for embedding and pooling methods. AvgPooling and MaxPooling are employed on the output of a specific layer to produce document embeddings.
%Here, we adopt Mask All for MDERank.
}
\label{table_pooling}
\end{table}

\begin{table*}[]
\centering
\resizebox{\textwidth}{!}{%
\begin{tabular}{c|ccc|ccc|ccc|ccc}
\hline
\multirow{2}{*}{\textbf{Method}} &
  \multicolumn{3}{c|}{\textbf{NUS $\left(512\right)$}} &
  \multicolumn{3}{c|}{\textbf{NUS $\left(2000\right)$}} &
  \multicolumn{3}{c|}{\textbf{Krapivin $\left(512\right)$}} &
  \multicolumn{3}{c}{\textbf{Krapivin $\left(2500\right)$}} \\ \cline{2-13} 
                        & F$_1$@5 & F$_1$@10 & F$_1$@15 & F$_1$@5 & F$_1$@10 & F$_1$@15 & F$_1$@5 & F$_1$@10 & F$_1$@15 & F$_1$@5 & F$_1$@10 & F$_1$@15 \\ \hline
EmbedRank(BERT)         & 3.75  & 6.34  & 8.11  & $-$   & $-$   & $-$   & 4.05  & 6.60  & 7.84  & $-$   & $-$   & $-$   \\
EmbedRank(BigBird) & 2.56  & 5.16  & 7.11  & 1.08  & 1.36  & 2.20  & 3.24  & 5.14  & 6.31  & 1.05  & 1.93  & 2.28  \\ \hline
MDERank(BERT)                 & 15.24 & 18.33 & 17.95 & $-$   & $-$   & $-$   & 11.78 & 12.93 & 12.58 & $-$   & $-$   & $-$   \\
MDERank(BigBird)         & 15.42 & 17.68 & 17.81 & 15.36 & 19.56 & 20.33 & 11.62 & 11.99 & 11.70 & 11.33 & 12.71 & 12.70 \\ \hline
\end{tabular}%
}
\caption{KPE performance from EmbedRank and MDERank using BERT and BigBird for embedding. 512, 2000, 2500 in the parentheses represent the number of words kept for each document in datasets. The results for NUS$\left(2000\right)$ and Krapivin $\left(2500\right)$ are missing for EmbedRank(BERT) and MDERank(BERT) due to limitation on input sequence length from BERT.}
\label{table_long}
\end{table*}

In the second experiment, we investigate effects of document length beyond 512 on EmbedRank and MDERank. To accommodate documents longer than 512, we choose BigBird~\cite{zaheer2020big} as the embedding model. BigBird replaces the full self-attention in Transformer with sparse attentions of global, local, and random attentions, reducing the quadratic complexity to sequence length from Transformer to linear. In order to select valid datasets for this evaluation, we count the average percentage of gold label KPs appearing in the first $m$ words in a document on the three longest datasets, DUC2001, NUS, and Krapivin. We observe that the first 500 words nearly cover 90\% gold KPs in DUC2001, whereas 50\% gold KPs in Krapivin are in the first 2500 words, and 50\% gold KPs in NUS are in the first 2000 words.
% As shown in Figure~\ref{Fig.main}, first 500 words nearly cover 90\% gold keyphrases in DUC2001, whereas 50\% gold keyphrases in Krapivin are concentrated in the first 2500 words, and 50\% gold KPs in NUS are concentrated in the first 2000 words.
Therefore, we drop DUC2001 and use NUS and Krapivin for the second experiment. We keep the first 2500 and 2000 words for documents in Krapivin and NUS, respectively.  Table~\ref{table_long} shows that on NUS, when increasing the document length from 512 to 2000, MDERank(BigBird) outperforms MDERank(BERT) by 2.38
%2.52
F$_1$@15. On Krapivin, when increasing the document length from 512 to 2500, 
MDERank(BigBird) also improves MDERank(BERT) by 0.12 F$_1$@15.
%MDERank also improves by 1.0 F$_1$@15.   
In contrast, the performance of EmbedRank degrades dramatically with longer context, since more context introduces more candidates into ranking and also worsens  
%more candidates will be involved in ranking, and 
the discrepancy between lengths of document and phrases, which in turn greatly reduces the accuracy of similarity comparison. \par

\begin{table*}[!htb]
\centering
\scalebox{0.85}{%
\begin{tabular}{c|c|ccccccc}
\hline
\multirow{2}{*}{\textbf{F$_1$@K}} & \multirow{2}{*}{\textbf{$\theta$}} & \multicolumn{7}{c}{\textbf{Dataset}}                                       \\ \cline{3-9} 
 &  & \textbf{Inspec} & \textbf{SemEval2017} & \textbf{SemEval2010} & \textbf{DUC2001} & \textbf{Krapivin} & \multicolumn{1}{c|}{\textbf{NUS}} & \textbf{AVG} \\ \hline
\multirow{2}{*}{5}              & TextRank                         & 28.93 & 21.34 & 11.46 & 13.30 & 7.85  & \multicolumn{1}{c|}{7.57}  & 15.08 \\
                                & YAKE                             & 28.06 & 21.63 & 12.95 & 22.51 & 12.91 & \multicolumn{1}{c|}{14.11} & 18.70 \\ \hline
\multirow{2}{*}{10}             & TextRank                         & 38.13 & 32.71 & 17.23 & 19.15 & 10.47 & \multicolumn{1}{c|}{10.59} & 21.38 \\
                                & YAKE                             & 35.80 & 32.23 & 17.95 & 26.97 & 14.36 & \multicolumn{1}{c|}{17.72} & 24.17 \\ \hline
\multirow{2}{*}{15}             & TextRank                         & 39.49 & 37.95 & 19.89 & 22.11 & 11.40 & \multicolumn{1}{c|}{12.83} & 23.95 \\
                                & YAKE                             & 37.43 & 37.52 & 20.69 & 26.28 & 13.58 & \multicolumn{1}{c|}{17.95} & 25.58 \\ \hline
\end{tabular}%
}
\caption{The KPE performance (F$_1$@K) from MDERank(KPEBERT) with KPEBERT pre-trained using YAKE and TextRank as $\theta$ for producing pseudo labels, respectively. \textbf{AVG} is the average F1@K on all six benchmarks}
\label{table_heuristic}
\end{table*}

\noindent \textbf{Effects of Encoder Layers and Pooling Methods}
The findings in~\citep{jawahar2019does,kim2020pre,rogers2020primer} show that BERT captures a rich hierarchy of linguistic information, with surface features in lower layers, syntactic features in middle layers, and semantic features in higher layers. We conduct experiments to understand the effects on MDERank and EmbedRank when using different BERT layers for embedding. We choose the third, the sixth, and the last layer from BERT-Base. We study the interactions between encoder layers and different Pooling methods. As shown in Table~\ref{table_pooling},  for both AvgPooling and MaxPooling, F$_1$ from MDERank(BERT) shows a steady gain to the increase of layers. On the contrary,  with AvgPooling, F$_1$ from EmbedRank(BERT) drastically drops as the layers rises from 3 to 12, probably due to that the lower BERT layer provides more rough and generic representations, which may alleviate mismatch in similarity comparison for Phrase-Document methods. We test the average F1@5, F1@10, F1@15 with the configuration for EmbedRank(BERT) that yields best results on DUC2001, i.e., AvgPooling and layer 3, on all 6 datasets and the results are 3.7, 1.8 and 1.6 absolute lower than MDERank(BERT). Compared to AvgPooling, MaxPooling produces weaker document embedding, which severely degrades the performance of EmbedRank and slightly degrades performance of MDERank. On the other hand, MaxPooling probably reduces differences in embeddings across layers, hence performance of EmbedRank becomes stable across layers with MaxPooling. 

For both pooling methods, MDERank using the last BERT layer achieves the best results, demonstrating that MDERank can fully benefit from stronger contextualized semantic representations.

\noindent \textbf{Effects of the Choice of $\theta$ on KPEBERT}
 We also investigate the effects of choosing different unsupervised KPE methods as $\theta$ for generating pseudo labels for KPEBERT pre-training. When balancing the extraction speed and KPE quality, TextRank is another choice for $\theta$ besides YAKE. As shown in Table~\ref{table_main}, YAKE performs better than TextRank on long-document datasets but worse on short-document datasets. After replacing YAKE with TextRank as $\theta$ for producing pseudo labels and training KPEBERT, the KPE results of the respective MDERank(KPEBERT) with absolute sampling are shown in Table~\ref{table_heuristic}. We observe that MDERank(KPEBERT) using YAKE as $\theta$ significantly outperforms MDERank(KPEBERT) using TextRank as $\theta$, on both short-document datasets and long-document datasets (except worse on Inspec and comparable on SemEval2017).  Although on average YAKE performs worse than TextRank on the six benchmarks, the better performance from YAKE on long documents coupled with its stable performance may be a crucial factor when choosing $\theta$ for pre-training KPEBERT. Results in Table~\ref{table_main} shows that MDERank(KPEBERT) with YAKE for pseudo labeling yields superior performance on both short and long documents. In other words, KPEBERT benefits from the stable performance from YAKE on long documents for pseudo labeling while exhibiting robustness to the relatively low performance on short documents from YAKE. 
 
\vspace{-1mm}
\section{Conclusion}
\vspace{-1mm}
We propose a novel embedding-based unsupervised KPE approach, MDERank, to improve reliability of similarity match compared to previous embedding-based methods. We also propose a novel self-supervised learning method and develop a KPE-oriented PLM, KPEBERT. Experiments demonstrate MDERank outperforms SOTA on diverse datasets and further benefits from KPEBERT. Analyses further verify the robustness of MDERank to different lengths of keyphrases and documents, and that MDERank benefits from longer context and stronger embedding models. Future work includes improving KPEBERT for MDERank by optimizing sampling strategies and pre-training methods.  

\section{Acknowledgements}
This work was supported by Alibaba Group through Alibaba Research Intern Program and A*STAR AME Programmatic Funding Scheme (Project No. A18A1b0045).

\bibliography{anthology}
\bibliographystyle{acl_natbib}

\clearpage
\appendix
\appendixpage
\section{Effects of Masking Methods on MDERank}
\label{appendix:masking}

\begin{table*}[]
\centering
\scalebox{0.85}{%
\begin{tabular}{c|c|ccccccl}
\hline
\multicolumn{1}{l|}{\multirow{2}{*}{\textbf{F$_1$@K}}} & \multirow{2}{*}{\textbf{Method}} & \multicolumn{7}{c}{\textbf{Dataset}}                                       \\ \cline{3-9} 
\multicolumn{1}{l|}{} &
   &
  \textbf{Inspec} &
  \textbf{SemEval2017} &
  \textbf{SemEval2010} &
  \textbf{DUC2001} &
  \textbf{Krapivin} &
  \multicolumn{1}{c|}{\textbf{NUS}} &
  \textbf{AVG} \\ \hline
\multirow{4}{*}{5}                                   & Mask All                         & 26.17 & 22.81 & 12.95 & 13.05 & 11.78 & \multicolumn{1}{c|}{15.24} & 17.00 \\
                                                     & Mask Once                        & 27.93 & 20.56 & 10.16 & 9.11  & 4.61  & \multicolumn{1}{c|}{3.92}  & 12.72 \\
                                                     & Mask Highest                     & 27.93 & 20.56 & 10.16 & 9.11  & 4.65  & \multicolumn{1}{c|}{3.92}  & 12.72 \\
                                                     & Mask Subset                      & 29.25 & 21.50 & 10.26 & 12.05 & 8.50  & \multicolumn{1}{c|}{9.61}  & 15.20 \\ \hline
\multirow{4}{*}{10}                                  & Mask All                         & 33.81 & 32.51 & 17.07 & 17.31 & 12.93 & \multicolumn{1}{c|}{18.33} & 21.99 \\
                                                     & Mask Once                        & 37.38 & 30.95 & 15.40 & 13.49 & 7.21  & \multicolumn{1}{c|}{6.52}  & 18.49 \\
                                                     & Mask Highest                     & 37.42 & 30.97 & 15.32 & 13.46 & 7.24  & \multicolumn{1}{c|}{6.56}  & 18.50 \\
                                                     & Mask Subset                      & 36.55 & 31.30 & 15.88 & 16.73 & 9.99  & \multicolumn{1}{c|}{13.43} & 20.65 \\ \hline
\multirow{4}{*}{15}                                  & Mask All                         & 36.17 & 37.18 & 20.09 & 19.13 & 12.58 & \multicolumn{1}{c|}{17.95} & 23.85 \\
                                                     & Mask Once                        & 39.11 & 36.07 & 17.69 & 16.47 & 8.15  & \multicolumn{1}{c|}{8.85}  & 21.06 \\
                                                     & Mask Highest                     & 39.36 & 36.10 & 17.76 & 16.45 & 8.20  & \multicolumn{1}{c|}{8.85}  & 21.12 \\
                                                     & Mask Subset                      & 38.08 & 36.67 & 17.83 & 19.19 & 10.48 & \multicolumn{1}{c|}{14.65} & 22.82 \\ \hline
\end{tabular}%
}
\caption{F$_1$@K ($K \in \{5, 10,15\}$) from MDERank(BERT) using different masking methods, where Mask All refers to the masking method described in Section~\ref{sec:mderank}.}
\label{table_variants}
\end{table*}

Given occurrences of a candidate KP $c_i$ in a document $d$ as $[p_1, p_2, \ldots, p_t]$, we study several methods to mask these occurrences and generate the masked document $d_M^{c_i}$, 
%For a candidate phrase, denote its occurrences in a document $d$ as $[p_1, p_2, \ldots, p_t]$. We have several strategies to create the masked document $d_M$, 
considering the potential bias e.g., frequency, sequence length, and nested phrases.

\textbf{Mask Once} The \emph{Mask Once} method only masks the first occurrence of a candidate.
This strategy eliminates the bias towards high frequency candidate KPs.
However, it may prefer longer candidate KPs (i.e., candidate KPs that
consist of more subwords) with the same argument shown in Section~\ref{sec:intro}. 
MDERank may benefit from this masking strategy on datasets with annotation bias towards long keyphrases. 

\textbf{Mask Highest}
The \emph{Mask Highest} method considers the collection of $d_M^{c_i}$s obtained by masking each occurrence
of a candidate phrase $c_i$ once in the document, and select the one that has the
\emph{smallest} cosine similarity with the embeddings of $d$. This method considers a balance of impacts from sequence length and frequency of candidate phrases.

\textbf{Mask Subset} One issue in KPE is that there may be heavy nesting
among candidate KPs. For example, ``support vector machine'' may result in nested
candidates such as ``support vector machine'', ``support vector'',
``vector machine'', and even ``machine''.  Neither \emph{Mask All} nor \emph{Mask Once} strategy addresses this issue and hence the nested KPs may take up a large proportion in the final results, drastically damaging the diversity. We design the \emph{Mask Subset} method to alleviate impact of nested candidate KPs. Firstly, all candidates are ranked by their phrase length in a descending order. Secondly, when generating a masked document for each candidate in order, \emph{Mask Subset} records the positions of masked words and requires that each candidate could only be masked with words not in the recorded positions.

\begin{figure}[htp]
\centering % 图片居中
\includegraphics[width=0.4\textwidth]{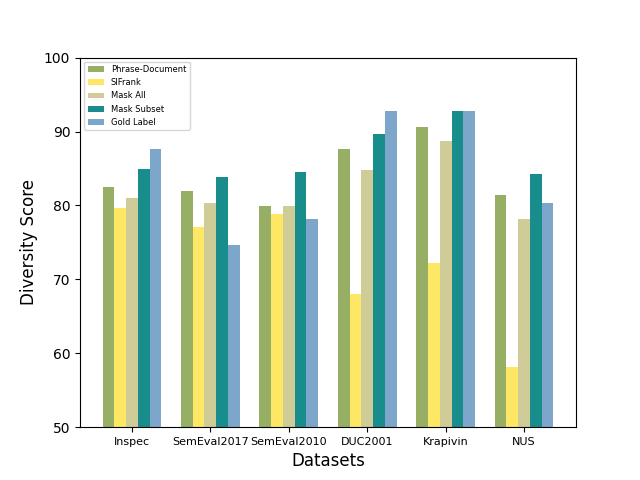}
\caption{Diversity scores from different methods on various datasets. A higher bar indicates a better diversity. The diversity of gold keyphrases are in blue and on the right. }
\label{figure_div}
\end{figure}

The KPE results from MDERank(BERT) using these masking strategies are shown in Table~\ref{table_variants}. The masking variants do not bring remarkable improvement compared with the results from Mask All, and Mask Once and Mask Highest perform even worse on the long-document datasets. This is because masking only one occurrence of a candidate will not emphasize the change of semantics significantly, especially on long documents.   
Mask subset could partially address the diversity problem by reducing the number of nested candidates selected by MDERank. Figure~\ref{figure_div} shows a comparison on diversity between \emph{Mask Subset} and other methods, where the evaluation metric for diversity is defined in Equation~\ref{eq:div}. The Phrase-Document method refers to EmbedRank(BERT).  We could see from Figure~\ref{figure_div} that MDERank with Mask Subset indeed boosts the diversity over Mask All and even exceeds gold labels on several datasets.
%As can be seen in Figure~\ref{figure_div}, MDERank(BERT) with Mask All indeed boost the diversity and even exceed gold labels on several datasets.

\begin{equation}
\label{eq:div}
      Diveristy(d)=  \frac{t_u}{t_n} *100
\end{equation}

\begin{table*}[!htb]
\centering
\scalebox{0.75}{%
\begin{tabular}{@{}c|c|ccccccc@{}}
\toprule
\multirow{2}{*}{\textbf{Method}} &
  \multirow{2}{*}{\textbf{F1@K}} &
  \multicolumn{7}{c}{\textbf{Dataset}} \\ \cmidrule(l){3-9} 
 &
   &
  \textbf{Inspec} &
  \textbf{SemEval2017} &
  \textbf{SemEval2010} &
  \textbf{DUC2001} &
  \textbf{Krapivin} &
  \textbf{NUS} &
  \textbf{AVG} \\ \midrule
\multirow{3}{*}{EmbedRank(Cos)} & 5  & 28.92 & 20.03 & 10.46 & 8.12  & 4.05  & 3.75           & 12.56 \\
                                & 10 & 38.55 & 31.01 & 16.35 & 11.62 & 6.60  & 6.34           & 18.41 \\
                                & 15 & 39.77 & 36.72 & 19.35 & 13.58 & 7.84  & 8.11           & 20.90 \\ \midrule
\multirow{3}{*}{EmbedRank(Euc)} & 5  & 29.28 & 19.77 & 9.47  & 7.92  & 4.13  & 4.04           & 12.44 \\
                                & 10 & 38.23 & 30.58 & 16.35 & 11.61 & 6.66  & 6.52           & 18.33 \\
                                & 15 & 39.80 & 36.14 & 19.02 & 13.49 & 7.71  & \textbf{8.18}  & 20.72 \\ \midrule
\multirow{3}{*}{MDERank(Cos)}   & 5  & 26.17 & 22.81 & 12.95 & 13.05 & 11.78 & 15.07          & 16.97 \\
                                & 10 & 33.81 & 32.51 & 17.07 & 17.31 & 12.93 & 19.20          & 22.14 \\
                                & 15 & 36.17 & 37.18 & 19.02 & 19.13 & 12.58 & \textbf{19.62} & 23.95 \\ \midrule
\multirow{3}{*}{MDERank(Euc)}   & 5  & 26.25 & 22.83 & 12.76 & 13.10 & 11.29 & 15.24          & 16.91 \\
                                & 10 & 33.83 & 32.59 & 17.15 & 17.45 & 12.15 & 18.29          & 21.91 \\
                                & 15 & 36.25 & 37.24 & 20.22 & 19.33 & 11.82 & 18.02          & 23.81 \\ \bottomrule
\end{tabular}%
}
\caption{The KPE performance from MDERank and EmbedRank using Cosine and Euclidean as similarity measure, where EmbedRank is EmbedRank(BERT) as in Section~\ref{sec:baseline} and MDERank is MDERank(BERT).}
\label{table_similarity}
\end{table*}

 %the extraction speed and quality, we replace YAKE by TextRank to rank candidates and the results are shown in Table~\ref{table_heuristic}. The experiment results demonstrate that KPEBERT based on YAKE performs more stable on diverse datasets and achieves better performance on long-document datasets. According to Table~\ref{table_main}, YAKE performs better on long-document dataset than TextRank but weaker on short-document datasets. It is noted that KPEBERT produces competitive performance not only on short documents but also on long documents. In other words, KPEBERT gain from YAKE results on long documents while not be affected on short documents. 

\section{Impact of Similarity Measure}
\label{appendix:similarity-measure}
The common similarity measures include Cosine and Euclidean distance. However, the choice of similarity measure does not matter for MDERank performance. We conduct experiments to investigate the impact of the similarity measure on the performance of MDERank, and the results are shown in Table~\ref{table_similarity}. We observe that Cosine and Euclidean similarity measure are not a salient factor for the ranking results for both EmbedRank(BERT) and MDERank(BERT).
%SIFRank and EmbedRank(BERT).

\end{document}